\documentclass[10pt,twocolumn,letterpaper]{article}
\usepackage{times}

\usepackage{xspace}
\usepackage[dvipsnames]{xcolor}
\usepackage{graphicx}
\usepackage{amsmath}
\usepackage{amssymb}
\usepackage{booktabs}
\usepackage[numbers,sort&compress]{natbib}
\usepackage{etoolbox}
\usepackage[format=plain,labelformat=simple,labelsep=period,font=small,compatibility=false]{caption}
\usepackage[font=footnotesize,skip=3pt,subrefformat=parens]{subcaption}
\usepackage[hyphens]{url}
\IfFileExists{latexml.sty}{\usepackage{latexml}}{\newif\iflatexml\latexmlfalse}
\usepackage[shortlabels,inline]{enumitem}
\setlist[itemize]{noitemsep,leftmargin=*,topsep=0em}
\setlist[enumerate]{noitemsep,leftmargin=*,topsep=0em}

\usepackage[letterpaper,textwidth=6.875in,textheight=8.875in,columnsep=0.3125in,centering]{geometry}
\usepackage{titlesec}
\titleformat{\section}{\large\bfseries}{\thesection.}{0.5em}{}
\titleformat{\subsection}{\normalsize\bfseries}{\thesubsection.}{0.5em}{}
\titlespacing*{\section}{0pt}{10pt plus 2pt minus 2pt}{7pt}
\titlespacing*{\subsection}{0pt}{8pt plus 2pt minus 2pt}{6pt}
\titleformat{\paragraph}[runin]{\normalsize\bfseries}{}{0pt}{}
\titlespacing*{\paragraph}{0pt}{6pt plus 2pt minus 1pt}{0.5em}
\iflatexml\else
\renewenvironment{abstract}{\centerline{\large\bfseries Abstract}\vspace{8pt}\noindent\itshape\ignorespaces}{\par\vspace{10pt}}
\fi

\makeatletter
\DeclareRobustCommand\onedot{\futurelet\@let@token\@onedot}
\def\@onedot{\ifx\@let@token.\else.\null\fi\xspace}
\def\eg{\emph{e.g}\onedot}

\def\etal{\emph{et al}\onedot}
\makeatother

\usepackage{microtype}

\usepackage{multirow}
\usepackage{colortbl}
\definecolor{linkblue}{rgb}{0.21,0.49,0.74}
\usepackage[pagebackref,breaklinks,colorlinks,allcolors=linkblue]{hyperref}

\usepackage[capitalize]{cleveref}
\crefname{section}{Sec.}{Secs.}
\Crefname{section}{Section}{Sections}
\Crefname{table}{Table}{Tables}
\crefname{table}{Tab.}{Tabs.}

\title{ADAPT: Adapting Vision Foundation Models to Dermatology on a Single GPU}

\iflatexml
  \usepackage{authblk}

  \author{Hasan Can Sa\u{g}lam†}
  \affil{Faculty of Engineering and Natural Sciences, Sabanc{\i} University, Istanbul, T\"{u}rkiye
    (\href{mailto:hasan.saglam@sabanciuniv.edu}{hasan.saglam@sabanciuniv.edu})}

  \author{Mustafa \"{O}zen†}
  \affil{Faculty of Engineering and Natural Sciences, Sabanc{\i} University, Istanbul, T\"{u}rkiye
    (\href{mailto:mustafa.ozen@sabanciuniv.edu}{mustafa.ozen@sabanciuniv.edu})}

  \author{Berrin Yanikoglu}
  \affil{Faculty of Engineering and Natural Sciences, Sabanc{\i} University, Istanbul, T\"{u}rkiye
    (\href{mailto:berrin@sabanciuniv.edu}{berrin@sabanciuniv.edu})}

  \date{† Equal contribution}
\else
  \author{Hasan Can Sa\u{g}lam\textsuperscript{\dag}\qquad Mustafa \"{O}zen\textsuperscript{\dag}\qquad Berrin Yanikoglu\\[4pt]
  {\normalsize\itshape Faculty of Engineering and Natural Sciences, Sabanc{\i} University, Istanbul, T\"{u}rkiye}\\[2pt]
  {\normalsize\{hasan.saglam, mustafa.ozen, berrin\}@sabanciuniv.edu}\\[4pt]
  {\small \textsuperscript{\dag}Equal contribution}}
  \date{}
\fi
\iflatexml\else
\makeatletter
\def\@maketitle{\newpage\null\vskip .3in
  \begin{center}{\Large\bfseries \@title\par}\vskip 18pt
  {\large\lineskip .5em\begin{tabular}[t]{c}\@author\end{tabular}\par}\vskip 20pt\end{center}}
\makeatother
\fi
\begin{document}
\maketitle
\begin{abstract}
Models pretrained on natural images often transfer poorly to visually distinct domains, and adapting them with masked self-distillation is costly because of multi-crop training. 
We introduce ADAPT, an efficient self-supervised domain adaptation pretraining framework, using dermatology imaging as a challenging testbed for domain shift. 
ADAPT uses a SwinV2 teacher-student architecture: the student learns from masked and augmented views, and an EMA teacher provides stable targets from an unmasked view.
With only one global and one local crop and no projection heads, the objective combines masked token-level distillation with global representation alignment, enabling the model to learn from unlabeled dermoscopic and clinical images. 
Pretrained on a curated corpus of over 121,000 images from the ISIC Archive, ADAPT yields transferable representations. Compared with supervised ImageNet initialization and self-supervised baselines including SimMIM, iBOT, and DINOv3, ADAPT achieves the best average linear probing rank and fine-tuning Macro F1, and its representations transfer to lesion segmentation. It also reduces pretraining time by 4.4× relative to iBOT with the same backbone and 3.0× relative to DINOv3. 
Code will be made publicly available.
\end{abstract}
\section{Introduction}
\label{sec:intro}
Automated skin lesion analysis can support dermatological triage and diagnosis~\cite{esteva2017dermatologist,codella2018isic2017,codella2019isic2018}, yet models pretrained on natural images transfer poorly to this domain.
Visually similar skin lesions can carry different diagnoses, and their appearance varies with device, acquisition protocol, anatomical site, and patient populations~\cite{tschandl2018ham10000,hernandezperez2024bcn20000,daneshjou2022ddi}.
As a result, off-the-shelf foundation models often experience severe performance degradation when transferred to highly specialized clinical and dermoscopic datasets~\cite{kawahara2019derm7pt,tschandl2026milk10k,chiou2025midas}.

Most high-performing skin lesion classifiers are trained in a supervised manner, requiring large amounts of accurately annotated data ~\cite{esteva2017dermatologist,codella2019isic2018}, but expert annotations in dermatology are costly and time-consuming. Self-supervised learning (SSL) provides a natural alternative by exploiting unlabeled images to learn transferable visual representations before supervised fine-tuning~\cite{he2022mae,grill2020byol,simeoni2025dinov3}. This is especially attractive in dermatology~\cite{azizi2021big}, where large image collections exist, but only a subset may have reliable task-specific labels.

\begin{figure}[t]
    \centering
    \includegraphics[width=\linewidth]{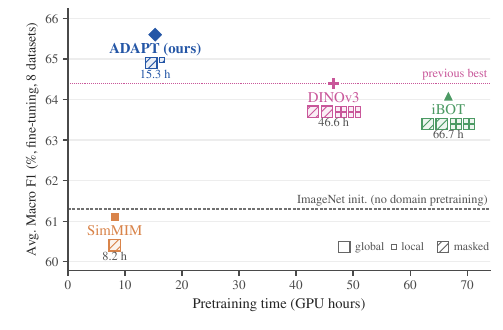}
    \caption{Average fine-tuning Macro F1 on 8 dermatology benchmarks vs.\ pretraining time (1$\times$ NVIDIA RTX PRO 6000). With two masked views (hatched glyphs), ADAPT surpasses DINOv3 by 1.2 points while pretraining 4.4$\times$ faster than iBOT.}
    \label{fig:teaser}
\end{figure}
Large dermatology foundation models, such as PanDerm~\cite{yan2025panderm}, show the value of domain-specific pretraining, but they require substantial data aggregation and computational resources. 
Continuing self-supervised pretraining from a general-purpose model on target-domain data~\cite{reed2022hpt} is a cheaper alternative. However, the strongest recipes for it, such as masked self-distillation methods like iBOT~\cite{zhou2022ibot} and DINOv3~\cite{simeoni2025dinov3}, rely on multi-crop training with ten or more views per image, which multiplies the pretraining cost. This motivates domain-adaptive SSL strategies that remain computationally practical and reproducible, using only a curated corpus from a single public dermatology archive.

We introduce ADAPT, a computationally efficient domain adaptation pretraining framework built on a SwinV2 backbone~\cite{liu2021swin,liu2022swinv2}.
%
Our main contributions are as follows:
\begin{itemize}
    \item \textbf{Simplified Multi-Crop Strategy:} Unlike iBOT~\cite{zhou2022ibot} and DINOv3~\cite{simeoni2025dinov3}, which rely on 
    multiple ($\ge 10$) crop configurations, ADAPT achieves  representation alignment using only a single global and a single local crop. 
 Masking both student views is what makes this reduced-crop setting effective; leaving the local crop unmasked reduces performance to the level of the ImageNet initialization.
    \item \textbf{Direct Dual-Level Distillation:} We eliminate projection heads by employing two lightweight student-side predictors, directly computing cosine similarity against the EMA teacher~\cite{grill2020byol} at the masked-token and global-image levels.

    \item \textbf{Domain Adaptation \& Efficiency:} We extensively evaluate ADAPT on dermatology images, a domain with large visual variability. Validated through linear probing, limited-label, and full fine-tuning evaluations on classification and segmentation, ADAPT achieves the best average downstream performance, particularly on dermoscopic benchmarks, while pretraining 4.4× faster than iBOT with the same backbone and 3.0× faster than DINOv3 (\cref{fig:teaser}).
    
\end{itemize}

\section{Related work}
\label{sec:related}

\paragraph{Skin lesion classification.}
Supervised deep networks have reached dermatologist-level accuracy~\cite{esteva2017dermatologist} and outperformed most of the 58 dermatologists in a dermoscopic reader study~\cite{haenssle2018man}; in the ISIC 2018 reader study, the best algorithms were more accurate than experienced dermatologists~\cite{tschandl2019comparison}. 
Challenge-winning systems, such as the top entry of ISIC 2019, typically rely on large supervised ensembles trained at multiple input resolutions~\cite{gessert2020skin}. 
These gains depend on expert labels, and performance degrades on images from unseen sources~\cite{tschandl2019comparison} and on underrepresented skin tones~\cite{groh2021fitzpatrick,daneshjou2022ddi}. 
Dermatology foundation models reduce the need for labels through large-scale pretraining: MONET~\cite{kim2024monet} aligns images with text from the medical literature, and PanDerm~\cite{yan2025panderm} applies self-supervised learning to a large multimodal corpus. ADAPT instead learns from unlabeled images of a single public archive on a single GPU.

\paragraph{
Self-supervised pretraining for medical images.}Self-supervised pretraining on unlabeled medical images outperforms supervised ImageNet initialization for dermatology and chest X-ray classification~\cite{azizi2021big}. Combined with large-scale supervised pretraining, it also improves robustness and data efficiency under distribution shift across six imaging domains~\cite{azizi2023remedis}. 
Masked autoencoders pretrained on the target dataset likewise improve medical image classification and segmentation~\cite{zhou2023selfpretraining}, and for skin lesions, Chaves \etal~\cite{chaves2022eval} compared self-supervised pipelines with a supervised baseline on in- and out-of-distribution data. Rather than pretraining from scratch, continuing self-supervised pretraining from an existing model on target-domain data converges faster and improves accuracy~\cite{reed2022hpt}. ADAPT follows this continued-pretraining strategy, starting from ImageNet-22k weights and adapting them to dermoscopic and clinical images.

\paragraph{Self-distillation and masked prediction.}
BYOL~\cite{grill2020byol} showed that a predictor and a momentum teacher enable learning without negative pairs, and SimSiam~\cite{chen2021simsiam} showed that a predictor with a stop-gradient suffices even without momentum.
Self-distillation methods such as DINO~\cite{caron2021dino} match teacher and student outputs through projection heads across multi-crop views~\cite{caron2020swav}, at a cost that grows with the number of crops; EsViT~\cite{li2022esvit} extends this to hierarchical Swin backbones with an additional region-level objective. 
Masked image modeling reconstructs the pixels of hidden patches~\cite{he2022mae,simmim}, whereas masked feature prediction~\cite{wei2022maskfeat} regresses a momentum teacher's representations of masked regions without image-level distillation or local crops, as in data2vec~\cite{baevski2022data2vec}, its more efficient successor data2vec~2.0~\cite{baevski2023data2vec2}, BootMAE~\cite{dong2022bootmae}, and I-JEPA~\cite{assran2023ijepa}; MSN~\cite{assran2022msn} and SiameseIM~\cite{tao2023siameseim} instead match a masked view to another view of the same image. Closest to ADAPT, iBOT~\cite{zhou2022ibot}, DINOv2~\cite{oquab2024dinov2} and DINOv3~\cite{simeoni2025dinov3} combine masked token-level and image-level self-distillation. ADAPT differs by using the teacher's backbone features directly as targets~\cite{liu2024dbot} instead of projection heads, by masking both student views, and by using only one global and one local crop.
\section{Methodology}

\begin{figure*}
    \centering
    \includegraphics[width=\textwidth]{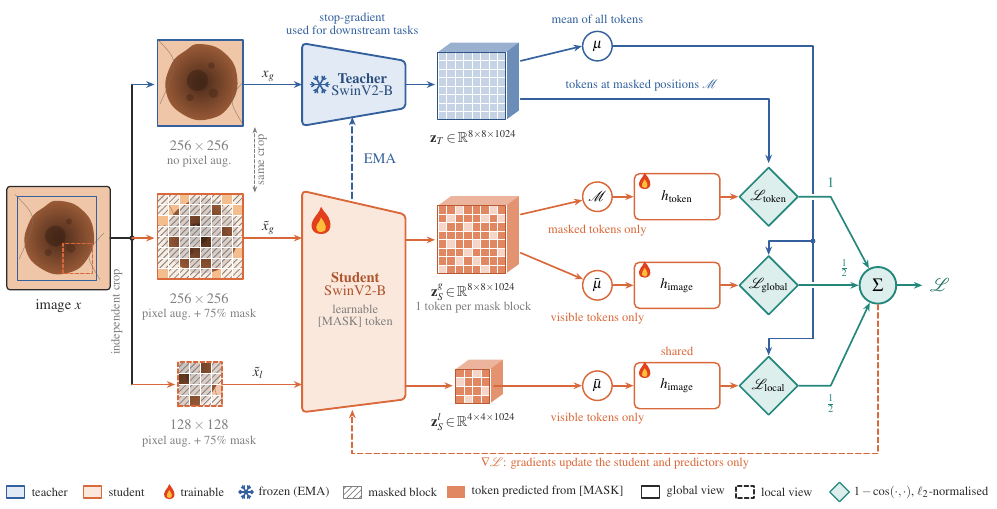}
    \caption{Overview of ADAPT. The frozen EMA teacher encodes a clean global crop, while the trainable student encodes masked versions of the same global crop and an independent local crop. Two predictors align student and teacher features at the masked-token and image levels.}
    \label{fig:method}
\end{figure*}

\subsection{Pretraining dataset collection}
To construct a diverse pretraining foundation, we aggregated all available clinical, including close-up and overview, and dermoscopic images from the ISIC Archive ~\cite{isic_archive}. To prevent data leakage and ensure rigorous downstream evaluation, we implemented a feature-based deduplication pipeline. We extracted image representations using a pretrained SwinV2-Base encoder and clustered the embeddings via DBSCAN~\cite{ester1996dbscan} using cosine distance with a strict similarity threshold of 0.975. If any image within a cluster intersected with the validation or test splits of our downstream datasets, the entire cluster was strictly excluded from the pretraining corpus. For the remaining duplicate clusters, only a single representative image was retained. Following this curation, our pretraining dataset consisted of 8,099 clinical images and 113,087 dermoscopic images.

\subsection{ADAPT pretraining framework}

\paragraph{Teacher-student architecture.}
ADAPT uses a teacher-student self-distillation framework,
following established self-distillation~\cite{grill2020byol} and masked self-distillation~\cite{zhou2022ibot}.
Both networks employ SwinV2-Base backbones (window size 16) initialized from the publicly released checkpoint pretrained on ImageNet-22k and fine-tuned on ImageNet-1k at $256 \times 256$~\cite{deng2009imagenet,liu2022swinv2}. The teacher processes a single unmasked global crop, while the student receives exactly one local crop alongside a masked and augmented version of the teacher's global crop. Gradients update only the student and its predictor networks, whereas the teacher is updated without backpropagation via EMA~\cite{he2020moco,grill2020byol}: $\theta_T \leftarrow m\theta_T + (1-m)\theta_S$,
where $\theta_S$ and $\theta_T$ denote the weights of the student and teacher networks, respectively, and $m$ is the EMA momentum. The final teacher backbone is used downstream.

Training lasted 100 epochs with AdamW~\cite{loshchilov2019adamw}, utilizing a 512 effective batch size, 0.05 weight decay, and 5.0 gradient clipping. Peak learning rates were $8 \times 10^{-4}$ for the backbone and $1 \times 10^{-3}$ for predictors under a 10\% warmup cosine schedule~\cite{loshchilov2017sgdr}. Layer-wise learning rate decay~\cite{clark2020electra,bao2022beit} was 0.9, and EMA momentum cosine-decayed from 0.999 to 1.0. We applied a 0.75 masking ratio with $32 \times 32$ patches across $256 \times 256$ global and $128 \times 128$ local crops.

\paragraph{Multi-crop augmentation strategy.}
We use a multi-crop augmentation strategy~\cite{caron2020swav}, extracting exactly one global and one local crop per image, to expose the student to both broad and detailed lesion views. 
The global crop uses a scale range of $0.4$--$1.0$ at $256 \times 256$ resolution, while the local crop uses a scale range of $0.05$--$0.4$ at $128 \times 128$ resolution, with both utilizing a random aspect ratio between $3/4$ and $4/3$. The crops undergo random transposition, horizontal and vertical flips, and shift-scale-rotate operations; the teacher and student global crops share the same spatial transformation, while the local crop is transformed independently. 

The teacher receives the corresponding \textit{unmasked} global crop. The student receives masked versions of both crops (with 75\% of the patches masked) alongside stronger pixel-level augmentations, including coarse dropout, CLAHE, HSV color jitter, brightness-contrast perturbation, and Gaussian blur, all implemented with Albumentations~\cite{buslaev2020albumentations}. 
This augmentation pipeline, both spatial and pixel-level, was selected based on preliminary experiments, where it yielded the most stable training. 
The chosen augmentations reflect typical variations in standard imaging conditions while simultaneously capturing the specific artifacts of our dermatology testbed, such as illumination, color, and hair occlusions.

\paragraph{Predictor and distillation objectives.}
Two student-side MLP predictors~\cite{grill2020byol,chen2021simsiam}, $h_{\mathrm{token}}$ and $h_{\mathrm{image}}$, are used for token-level and image-level distillation (\cref{fig:method}); both are two-layer MLPs ($1024 \to 2048 \to 1024$) with LayerNorm and GELU. All comparisons use cosine distance, $d(a,b)=1-\cos(a,b)$. The final-stage tokens of the backbone $f$ are
\begin{equation}
\mathbf{z}_T = f_{\theta_T}(x_g), \quad \mathbf{z}^{g}_S = f_{\theta_S}(\tilde{x}_g), \quad \mathbf{z}^{l}_S = f_{\theta_S}(\tilde{x}_l),
\label{eq:tokens}
\end{equation}
where $x_g$ is the teacher's global crop, $\tilde{x}_g$ its masked, augmented student copy, and $\tilde{x}_l$ the masked local crop.

\paragraph{Token-level distillation.} Similar to masked feature prediction in iBOT~\cite{zhou2022ibot}, data2vec~\cite{baevski2022data2vec}, and I-JEPA~\cite{assran2023ijepa}, $h_{\mathrm{token}}$ aligns student and teacher tokens at the masked positions $\mathcal{M}$ of the global crop, using a mask downsampled to the feature resolution:
\begin{equation}
\mathcal{L}_{\mathrm{token}} = \frac{1}{|\mathcal{M}|}\sum_{i\in\mathcal{M}} d\big(h_{\mathrm{token}}(\mathbf{z}^{g}_{S,i}),\, \mathbf{z}_{T,i}\big),
\label{eq:token}
\end{equation}
where $i$ indexes token positions. 
%

\paragraph{Image-level distillation.} Extending the global alignment of BYOL~\cite{grill2020byol} to masked representations, each masked student crop $v \in \{g, l\}$ is summarized by averaging its visible token positions $\mathcal{V}_v$, and the teacher's global representation by averaging all $N$ of its tokens:
\begin{equation}
\bar{\mathbf{z}}^{v}_S = \frac{1}{|\mathcal{V}_v|}\sum_{i\in\mathcal{V}_v}\mathbf{z}^{v}_{S,i}, \qquad
\bar{\mathbf{z}}_T = \frac{1}{N}\sum_{i=1}^{N}\mathbf{z}_{T,i}.
\label{eq:pool}
\end{equation}
Both student summaries are independently passed through $h_{\mathrm{image}}$ and aligned with the teacher:
\begin{equation}
\begin{aligned}
\mathcal{L}_{\mathrm{global}} &= d\big(h_{\mathrm{image}}(\bar{\mathbf{z}}^{g}_{S}),\, \bar{\mathbf{z}}_{T}\big),\\
\mathcal{L}_{\mathrm{local}} &= d\big(h_{\mathrm{image}}(\bar{\mathbf{z}}^{l}_{S}),\, \bar{\mathbf{z}}_{T}\big).
\end{aligned}
\label{eq:image}
\end{equation}

Whereas the token-level loss learns local structure, the image-level losses require the student to recover the teacher’s summary of the full, clean image from a partial, corrupted view. This encourages image-level representations that are invariant to masking, appearance changes and, for the local crop, scale and position. Pooling only visible tokens keeps the [MASK] outputs, which serve the token-level loss, out of the summary.

\paragraph{Total objective.} Each loss is averaged over the mini-batch to form the final training objective:
\begin{equation}
    \mathcal{L}_{\mathrm{ADAPT}} =
    \mathcal{L}_{\mathrm{token}} +
    \frac{1}{2}(\mathcal{L}_{\mathrm{global}} +
    \mathcal{L}_{\mathrm{local}}).
    \label{eq:loss}
\end{equation}

Since all terms are cosine distances on the same scale, the factor of 1/2 averages the two image-level terms so that token-level and image-level distillation contribute equally; we did not tune this weighting.
\section{Experimental setup}

Given the computational cost of large-scale self-supervised pretraining, exhaustive hyperparameter optimization was not performed. Instead, hyperparameters for ADAPT and the self-supervised baselines were selected through preliminary experiments on reduced subsets of the training data, in which multiple candidate values were manually evaluated. These trials favored conservative update dynamics, specifically a highly conservative EMA momentum, to ensure stable pretraining. Downstream fine-tuning hyperparameters were selected separately via manual evaluation on the task-specific validation splits to accommodate local adaptation.

\subsection{Baselines}

\begin{figure*}[t]
\centering
\includegraphics[width=\textwidth]{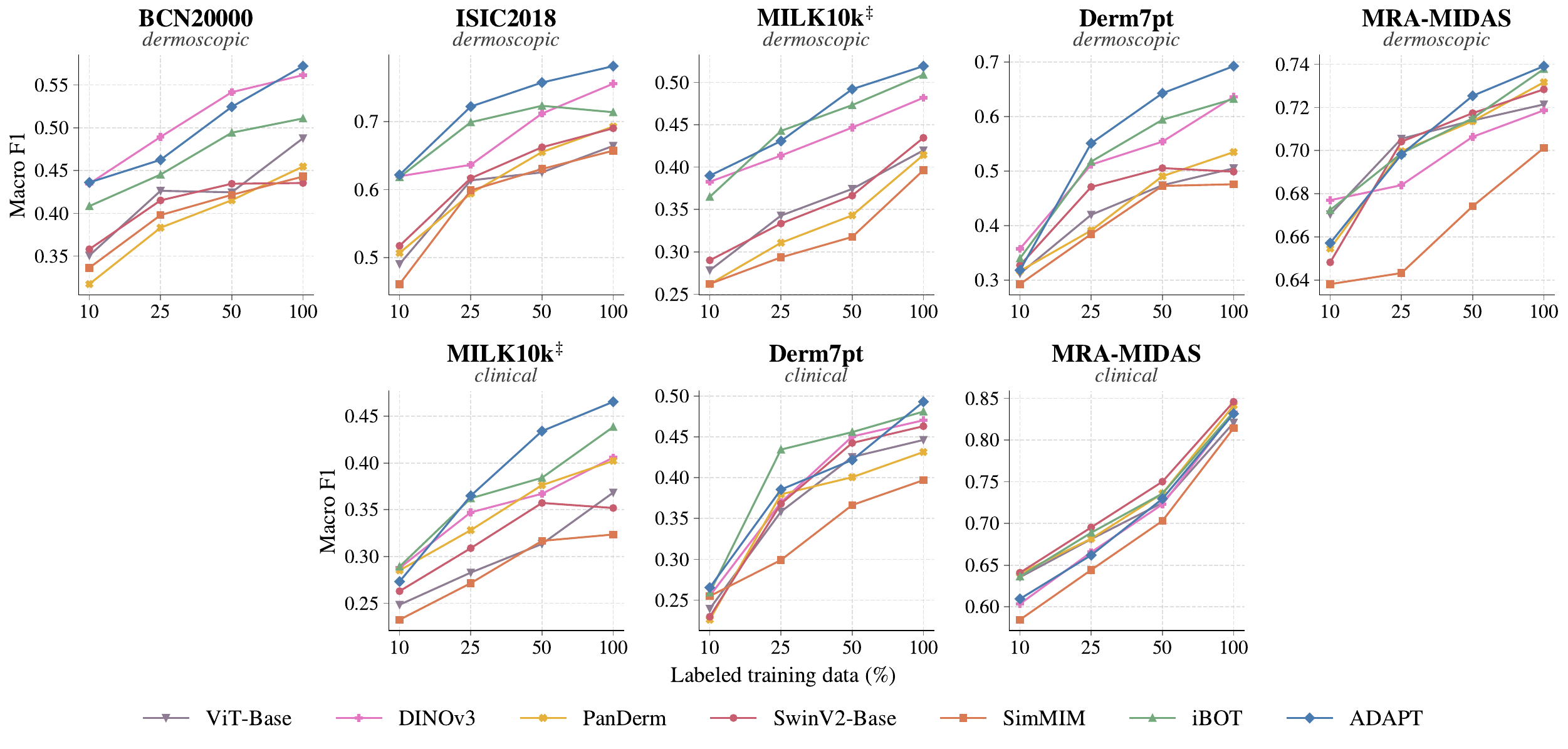}
\caption{Linear probing performance across varying labeled-data fractions 
(10\%, 25\%, 50\%, and 100\% of the training set) for dermoscopic and clinical classification datasets (Macro F1). 
Solid lines represent the mean across three seeds. A double dagger ($^{\ddagger}$) next to a dataset title indicates that results are 
reported on the validation split rather than the test set.}
\label{fig:linear_probing}
\end{figure*}

We compared ADAPT against supervised and self-supervised baselines under consistent downstream evaluation protocols, ensuring all models operate at the ``Base'' scale for fair comparison. \textbf{SwinV2-Base} denotes the supervised ImageNet-22k initialized base model using the same backbone as ADAPT.

To ensure a more controlled comparison, we performed 100 epochs of dermatology-specific self-supervised pretraining for the SSL baselines before downstream evaluation. \textbf{SimMIM}~\cite{simmim} and \textbf{iBOT}~\cite{zhou2022ibot} use the same SwinV2-Base backbone and ImageNet initialization as ADAPT, and share common pretraining settings (\eg, masking ratios and block sizes) unless otherwise noted. Specifically, SimMIM was trained using an adapted peak learning rate of $5 \times 10^{-5}$ and omitted the shift-scale-rotate operation and strong pixel-level student augmentations: coarse dropout, CLAHE, HSV color jitter, and Gaussian blur. iBOT follows a ten-crop configuration comprising two masked global and eight unmasked local views.

We additionally evaluated \textbf{DINOv3}~\cite{simeoni2025dinov3} using a ViT-Base~\cite{dosovitskiy2021vit} backbone (to ensure architectural consistency for the DINOv3 method), initialized from its publicly available pretrained weights, followed by 100 epochs of continued self-supervised pretraining on dermatology images. For this pretraining, DINOv3 utilizes the same hyperparameter settings, scheduling, and augmentations as proposed in its original paper, with the exception of using an effective batch size of 512, applying a layer-wise learning rate decay of 0.9, adapting the ten-crop configuration to use $256 \times 256$ global and $128 \times 128$ local crops, and conducting the high-resolution stage at a $512 \times 512$ resolution for both Gram and global images alongside $128 \times 128$ local crops with an effective batch size of 128. This allows us to assess the effectiveness of domain-specific pretraining across different self-supervised learning approaches, while acknowledging the architectural differences between ViT-Base and SwinV2-Base. \textbf{ViT-Base} denotes the same public DINOv3 weights fine-tuned directly, without dermatology pretraining. We also include \textbf{PanDerm}~\cite{yan2025panderm}, a large-scale dermatology foundation model, fine-tuned from its publicly released weights without additional pretraining on our data.

\subsection{Downstream datasets}
To ensure robust evaluation across modalities, datasets lacking official splits were divided into 60\% train, 20\% validation, and 20\% test partitions, meticulously ensuring images of the same physical lesion remained in the same split. For the MRA-MIDAS dataset, due to its smaller size, we instead performed 5-fold cross-validation (3 folds train, 1 fold validation, 1 fold test) and averaged the final metrics, maintaining this strict lesion-level partitioning across all folds.

\paragraph{Classification datasets.} We evaluate image-level transfer on five benchmarks:
\begin{itemize}
    \item \textbf{BCN20000:} Contains 9 classes of dermoscopic images exhibiting a long-tail distribution~\cite{hernandezperez2024bcn20000}. The training split (7,473 images) is used for pretraining, and the validation (2,495) and test (2,445 images) splits for evaluation.

    \item \textbf{ISIC2018 (Task 3):} Comprises 7 classes~\cite{codella2019isic2018,tschandl2018ham10000}. The official training split (10,015 images) is used for pretraining, while evaluation follows the official validation (193) and test (1,512 images) splits.

    \item \textbf{MILK10k:} A paired dataset of clinical and dermoscopic images with 11 long-tailed classes~\cite{tschandl2026milk10k,milkstudyteam2025milk10k}. The training split (4,097 pairs) is included in pretraining. Evaluation is performed separately on dermoscopic and clinical modalities using the official test set (479 pairs), while a validation split (1,143 pairs) is constructed from the training data. Linear probing results are reported on this validation split, whereas fine-tuning results are reported on the official benchmark test set due to submission constraints.

    \item \textbf{Derm7pt:} Contains paired dermoscopic and clinical images with official splits (413 training, 203 validation, and 395 test pairs)~\cite{kawahara2019derm7pt}. Evaluation is performed separately on dermoscopic and clinical modalities.

    \item \textbf{MRA-MIDAS:} We treat this dataset as a malignant detection benchmark, consisting of a curated subset of 908 unique, biopsy-proven benign and malignant lesions~\cite{chiou2025midas,stanfordaimi2024mramidas}. Because a single lesion can have multiple associated images, we grouped the dataset strictly by lesion. Evaluation is performed separately on dermoscopic and clinical modalities.
    
\end{itemize}
\paragraph{Segmentation datasets.} For dense prediction, we evaluate on two datasets:
\begin{itemize}
    \item \textbf{ISIC2018 (Task 1):} Lesion boundary segmentation with binary masks~\cite{codella2019isic2018}, using the official training (2,594), validation (100), and test (1,000 images) splits.

    \item \textbf{PH2:} 200 dermoscopic images with expert lesion masks~\cite{mendonca2013ph2}, used only as an external test set: models trained on ISIC2018 are tested on PH2 directly.
\end{itemize}

\subsection{Evaluation protocols}

\definecolor{oursrow}{gray}{0.93}
\newcommand{\res}[2]{$#1_{\pm #2}$}             
\newcommand{\fst}[2]{$\mathbf{#1}_{\pm #2}$}    
\newcommand{\snd}[2]{$\underline{#1}_{\pm #2}$} 

\begin{table*}
\centering
\caption{Classification fine-tuning results (Macro F1, mean$_{\pm\text{std}}$ over three seeds; sample standard deviation). Best in \textbf{bold}, second best \underline{underlined}.}
\label{tab:classification_finetuning}
\resizebox{\linewidth}{!}{%
\setlength{\tabcolsep}{4pt}
\renewcommand{\arraystretch}{1.15}
\begin{tabular}{@{}ll ccccc ccc c@{}}
\toprule
\multirow{2}{*}{\textbf{Method}} & \multirow{2}{*}{\textbf{Backbone}} &
\multicolumn{5}{c}{\textbf{Dermoscopic}} & \multicolumn{3}{c}{\textbf{Clinical}} & \multirow{2}{*}{\textbf{Avg.}} \\
\cmidrule(lr){3-7}\cmidrule(lr){8-10}
 & & BCN20000 & ISIC2018 & MILK10k & Derm7pt & MRA-MIDAS & MILK10k & Derm7pt & MRA-MIDAS & \\
\midrule
ViT-Base~\cite{simeoni2025dinov3} & \multirow{3}{*}{ViT-B}
 & \res{0.524}{.044} & \res{0.793}{.021} & \res{0.443}{.024} & \res{0.622}{.062} & \res{0.719}{.011}
 & \res{0.402}{.024} & \res{0.495}{.058} & \res{0.853}{.007} & 0.606 \\
DINOv3~\cite{simeoni2025dinov3} &
 & \res{0.577}{.017} & \res{0.823}{.006} & \snd{0.478}{.015} & \snd{0.724}{.009} & \snd{0.745}{.009}
 & \snd{0.405}{.036} & \snd{0.538}{.008} & \res{0.859}{.005} & \underline{0.644} \\
PanDerm~\cite{yan2025panderm} &
 & \res{0.506}{.011} & \res{0.769}{.027} & \res{0.436}{.021} & \res{0.610}{.036} & \res{0.725}{.007}
 & \res{0.351}{.023} & \res{0.513}{.067} & \snd{0.862}{.003} & 0.597 \\
\midrule
SwinV2-Base~\cite{liu2022swinv2} & \multirow{4}{*}{SwinV2-B}
 & \res{0.522}{.011} & \res{0.798}{.015} & \res{0.451}{.016} & \res{0.631}{.045} & \res{0.729}{.010}
 & \res{0.381}{.003} & \res{0.528}{.012} & \fst{0.864}{.004} & 0.613 \\
SimMIM~\cite{simmim} &
 & \res{0.536}{.020} & \res{0.780}{.010} & \res{0.448}{.016} & \res{0.679}{.017} & \res{0.718}{.012}
 & \res{0.383}{.019} & \res{0.489}{.038} & \res{0.853}{.004} & 0.611 \\
iBOT~\cite{zhou2022ibot} &
 & \snd{0.580}{.037} & \snd{0.835}{.005} & \res{0.471}{.011} & \res{0.717}{.021} & \res{0.742}{.011}
 & \res{0.395}{.006} & \res{0.528}{.035} & \snd{0.862}{.012} & 0.641 \\
\rowcolor{oursrow}
\textbf{ADAPT (Ours)} &
 & $\mathbf{0.589}_{\pm .019}$ & $\mathbf{0.852}_{\pm .012}$ & $\mathbf{0.502}_{\pm .027}$ & $\mathbf{0.740}_{\pm .012}$ & $\mathbf{0.754}_{\pm .009}$
 & \fst{0.407}{.038} & \fst{0.543}{.029} & \res{0.860}{.005} & \textbf{0.656} \\
\bottomrule
\end{tabular}%
}
\end{table*}

\paragraph{Metrics.}
For classification tasks, we utilized Macro F1 (the unweighted mean of class-wise F1 scores) to account for severe class imbalances.
For segmentation, we report the intersection over union (IoU) of the lesion class on the test set, computed over all test pixels pooled across images, with predictions binarized at a threshold of 0.5.

To ensure reproducibility and account for variability across experimental runs, all downstream evaluations were repeated using three fixed random seeds (42, 47, and 125). We report the mean score across these three runs, along with the standard deviation. 
To strictly evaluate the standalone performance of the models, test-time augmentation (TTA) was not employed during any inference phase.

\paragraph{Frozen-encoder evaluation.}
We evaluated the frozen encoder's representations by training a logistic regression classifier (scikit-learn~\cite{pedregosa2011sklearn}) using the L-BFGS solver with a maximum of 1000 iterations. The regularization strength was selected via grid search over $C \in \{10^{-3}, 10^{-2}, \dots, 10^{3}\}$ on the validation split. Features were extracted by forwarding unaugmented, resized, and normalized images through the frozen encoder exactly once.

For segmentation, feature extraction depends on the backbone architecture. For the hierarchical SwinV2 encoders, the outputs of the four stages are bilinearly upsampled to the resolution of the first stage and concatenated. For the isotropic ViT encoders, the tokens from the 3rd, 6th, 9th, and 12th (final) blocks are reshaped into 2D spatial grids and concatenated directly without upsampling, as they maintain a constant spatial resolution. In both cases, the concatenated features are mapped to a lesion mask by a single $1\times1$ convolution, trained for 50 epochs with AdamW and a combined binary cross-entropy and Dice loss; the checkpoint with the best validation IoU is evaluated.

To assess how efficiently the learned representations transfer when labels are limited, we repeated linear probing under several data regimes, fitting the classifier on 10\%, 25\%, 50\%, and 100\% subsets of the training data while always evaluating on the full held-out split. This isolates the effect of labeled-data quantity on downstream performance, reflecting dermatology settings where expert annotations are scarce.

\paragraph{Fine-tuning.}
For classification, models were fine-tuned for 50 epochs using cross-entropy loss, batch size 64, and inverse-frequency weighted sampling for class imbalance; the checkpoint with the best validation Macro F1 is evaluated on the test set. We employed a layer-wise learning rate decay of 0.9, with a base learning rate of $2 \times 10^{-4}$ for the backbone and a higher learning rate of $1 \times 10^{-3}$ for the linear classification head, both optimized using a cosine decay schedule with a 10\% warmup phase, a weight decay of 0.05, and gradient clipping at 5.0.

During fine-tuning, inputs were resized to $256 \times 256$ and augmented using the same spatial and pixel-level transformation pipeline as in the SSL local views, except that random resized cropping was replaced by fixed resizing.

To further improve generalization, we applied a hybrid MixUp~\cite{zhang2017mixup} and CutMix~\cite{yun2019cutmix} strategy with probability 0.25 for each. When activated, samples were paired within the batch and either interpolated (MixUp) or spatially mixed (CutMix) using Beta-distributed mixing coefficients $\alpha=0.8$ for MixUp and $\alpha=1.0$ for CutMix.
\section{Results}

\subsection{Linear probing performance}
\label{sec:linear_probing}

To evaluate the quality of the learned representations without updating the encoder weights, we first present the linear probing results for the classification task (\cref{fig:linear_probing}). Across all eight datasets and four label fractions, ADAPT achieves the best average rank among the seven models and ranks first in 19 of 32 settings, over twice as often.

On the dermoscopic datasets, ADAPT achieves top performance across all benchmarks in the full-label setting. This advantage is largely preserved in lower-data regimes, where ADAPT outperforms the baselines in 9 of the 15 limited-data scenarios. These results highlight the label efficiency of the learned representations.

On clinical close-up images, ADAPT achieves the best full-label results on two of the three benchmarks, MILK10k and Derm7pt. On the third benchmark, ADAPT remains strictly competitive with the top baselines, and it continues to demonstrate strong representational efficiency across the limited-data clinical regimes.

\subsection{Fine-tuning performance}

\Cref{tab:classification_finetuning} reports the full fine-tuning results. ADAPT achieves the highest Macro F1 score on seven of the eight benchmarks, yielding the best overall average (0.656, compared to 0.644 for DINOv3 and 0.641 for iBOT). On all five dermoscopic datasets, ADAPT consistently outperforms the baselines. For the clinical datasets, ADAPT achieves top performance on two of the three benchmarks. On the remaining clinical dataset, MRA-MIDAS, where all models fall within a narrow performance band (0.853--0.864), ADAPT remains within 0.004 of the best result.

\subsection{Segmentation performance}

\begin{figure}[t]
\centering
\includegraphics[width=\linewidth]{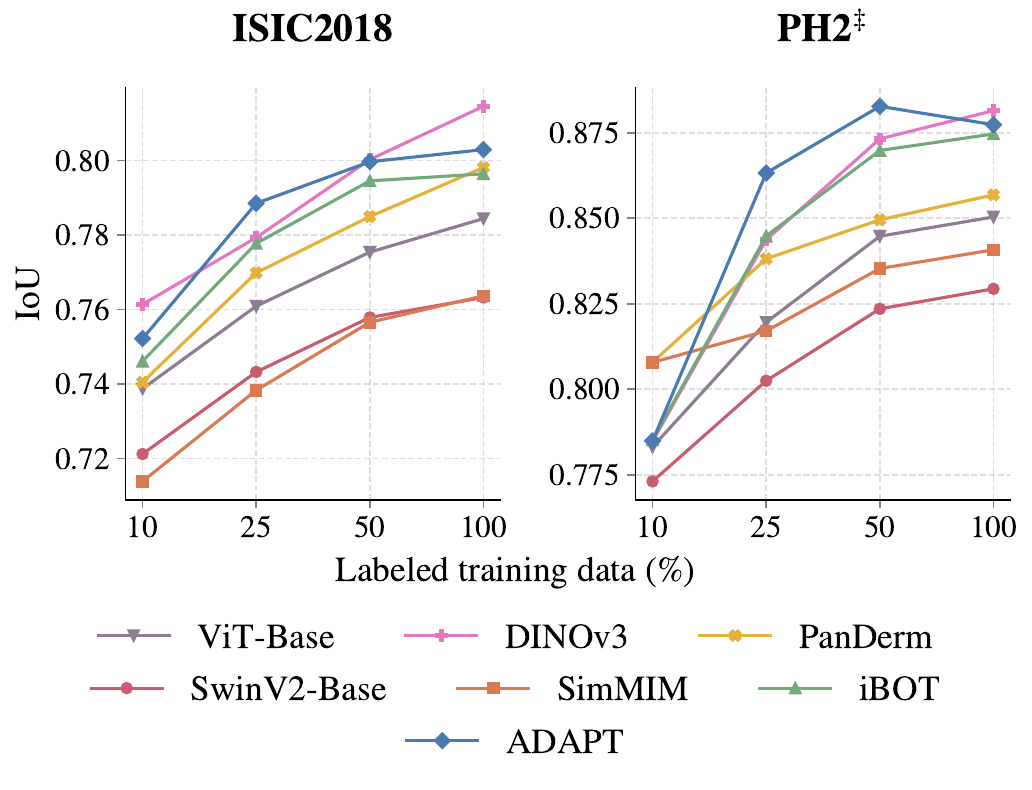}
\caption{Lesion segmentation with frozen encoders (linear probing, IoU) across labeled-data fractions (10\%, 25\%, 50\%, and 100\% of the training set). Solid lines represent the mean across three seeds. A double dagger ($^{\ddagger}$) marks PH2, an external test set: the probe is trained on ISIC2018 only.}
\label{fig:segmentation}
\end{figure}

\Cref{fig:segmentation} evaluates the frozen encoders on lesion segmentation. ADAPT achieves the highest IoU among all models on PH2 at 25\% and 50\% labels. While DINOv3, which uses a ViT-B backbone, remains strongest on ISIC2018 at 10\% and 100\% labels and on PH2 at 100\% labels, ADAPT maintains highly competitive performance in these regimes. These results indicate that ADAPT's representations transfer beyond image-level classification to dense prediction, consistent with its token-level distillation objective.

\subsection{Pretraining efficiency}

We compare the pretraining cost of each model on the same hardware, a single NVIDIA RTX PRO 6000 GPU. All times refer to one pretraining run (a single seed); each model was pretrained with three seeds. \cref{tab:efficiency} summarizes the results.

\begin{table}[!h]
\centering
\caption{Pretraining cost per run. Total time is the mean epoch time $\times$ 100 epochs. 
Avg. F1 is from 
\cref{tab:classification_finetuning}.}
\label{tab:efficiency}
\resizebox{\linewidth}{!}{
\setlength{\tabcolsep}{5pt}
\begin{tabular}{@{}llcccc@{}}
\toprule
\textbf{Method} & \textbf{Backbone} & \textbf{Crops/img} & \textbf{Time/epoch} & \textbf{Total} & \textbf{Avg.\ F1} \\
\midrule
SimMIM~\cite{simmim} & SwinV2-B & 1  & 4m 55s  & 8.2\,h  & 0.611 \\
iBOT~\cite{zhou2022ibot}   & SwinV2-B & 10 & 40m 00s & 66.7\,h & 0.641 \\
DINOv3~\cite{simeoni2025dinov3} & ViT-B    & 10 & 27m 57s & 46.6\,h & 0.644 \\
\rowcolor{oursrow}
\textbf{ADAPT (Ours)} & SwinV2-B & 2 & 9m 11s & \textbf{15.3\,h} & \textbf{0.656} \\
\bottomrule
\end{tabular}}
\end{table}

By using a single global and a single local crop, ADAPT trains 4.4$\times$ faster than iBOT and 3.0$\times$ faster than DINOv3, while also achieving a higher average fine-tuning score. SimMIM is cheaper, since it processes a single view without a teacher, but it trails ADAPT on every benchmark. ADAPT therefore offers the most favorable trade-off between pretraining cost and downstream performance.

\subsection{Ablation study}

To isolate each design choice, we first pretrain variants on a fixed random 10\% subset of the pretraining corpus for 100 epochs with three seeds and evaluate them by linear probing and fine-tuning on BCN20000 and ISIC2018 with 100\% labels. Because this schedule has about 10$\times$ fewer optimization steps, we lower the base EMA momentum from 0.999 to 0.99; all other hyperparameters are unchanged. The closest variants are then retrained on the full corpus and evaluated on all eight benchmarks (\cref{tab:ablation}).

\begin{table}[t]
\centering
\caption{Ablation study. Variants are first compared with 10\% of the pretraining corpus (on BCN20000 and ISIC2018); the closest ones are then retrained on the full corpus (on all eight benchmarks). Macro F1 averaged over benchmarks, mean$_{\pm\text{std}}$ over three seeds. Best in \textbf{bold}, second best \underline{underlined}. SwinV2-Base: ImageNet-22k-1k initialization without domain pretraining.}
\label{tab:ablation}
\resizebox{\linewidth}{!}{%
\setlength{\tabcolsep}{5pt}
\renewcommand{\arraystretch}{1.15}
\begin{tabular}{@{}l ccc r@{}}
\toprule
\textbf{Variant} & \textbf{Linear probing} & \textbf{Fine-tuning} & \textbf{Avg.} & \textbf{Time} \\
\midrule
\multicolumn{5}{c}{\textit{Pretrained on 10\% of the corpus}} \\[1pt]
SwinV2-Base~\cite{liu2022swinv2} & $0.562_{\pm .001}$ & $0.660_{\pm .013}$ & $0.611_{\pm .007}$ & -- \\
Global only, image-level & $\mathbf{0.621}_{\pm .030}$ & $0.668_{\pm .020}$ & $0.644_{\pm .025}$ & 1.1\,h \\
Global only, image + token & $0.607_{\pm .018}$ & $\mathbf{0.686}_{\pm .012}$ & $\underline{0.647}_{\pm .015}$ & 1.1\,h \\
ADAPT w. unmasked local crop & $0.555_{\pm .017}$ & $0.670_{\pm .017}$ & $0.612_{\pm .017}$ & 1.6\,h \\
ADAPT w. 25\% masked local crop & $0.574_{\pm .020}$ & $0.672_{\pm .012}$ & $0.623_{\pm .012}$ & 1.6\,h \\
ADAPT w. 50\% masked local crop & $0.578_{\pm .030}$ & $0.663_{\pm .010}$ & $0.620_{\pm .014}$ & 1.6\,h \\
ADAPT w. projection heads & $0.556_{\pm .014}$ & $0.665_{\pm .014}$ & $0.610_{\pm .014}$ & 1.6\,h \\
ADAPT w. more crops (2 global, 8 local) & $0.602_{\pm .015}$ & $0.670_{\pm .023}$ & $0.636_{\pm .019}$ & 6.1\,h \\
\rowcolor{oursrow}
\textbf{ADAPT (Ours)} & $\underline{0.615}_{\pm .023}$ & $\underline{0.681}_{\pm .017}$ & $\mathbf{0.648}_{\pm .020}$ & 1.6\,h \\
\midrule
\multicolumn{5}{c}{\textit{Pretrained on the full corpus}} \\[1pt]
SwinV2-Base~\cite{liu2022swinv2} & $0.556_{\pm .001}$ & $0.613_{\pm .015}$ & $0.585_{\pm .008}$ & -- \\
Global only, image-level & $\underline{0.634}_{\pm .011}$ & $\underline{0.648}_{\pm .014}$ & $\underline{0.641}_{\pm .012}$ & 11.0\,h \\
Global only, image + token & $0.629_{\pm .015}$ & $0.647_{\pm .015}$ & $0.638_{\pm .015}$ & 11.0\,h \\
\rowcolor{oursrow}
\textbf{ADAPT (Ours)} & $\mathbf{0.637}_{\pm .015}$ & $\mathbf{0.656}_{\pm .019}$ & $\mathbf{0.646}_{\pm .017}$ & 15.3\,h \\
\bottomrule
\end{tabular}}
\end{table}

\begin{figure*}[t]
\centering
\includegraphics[width=\textwidth]{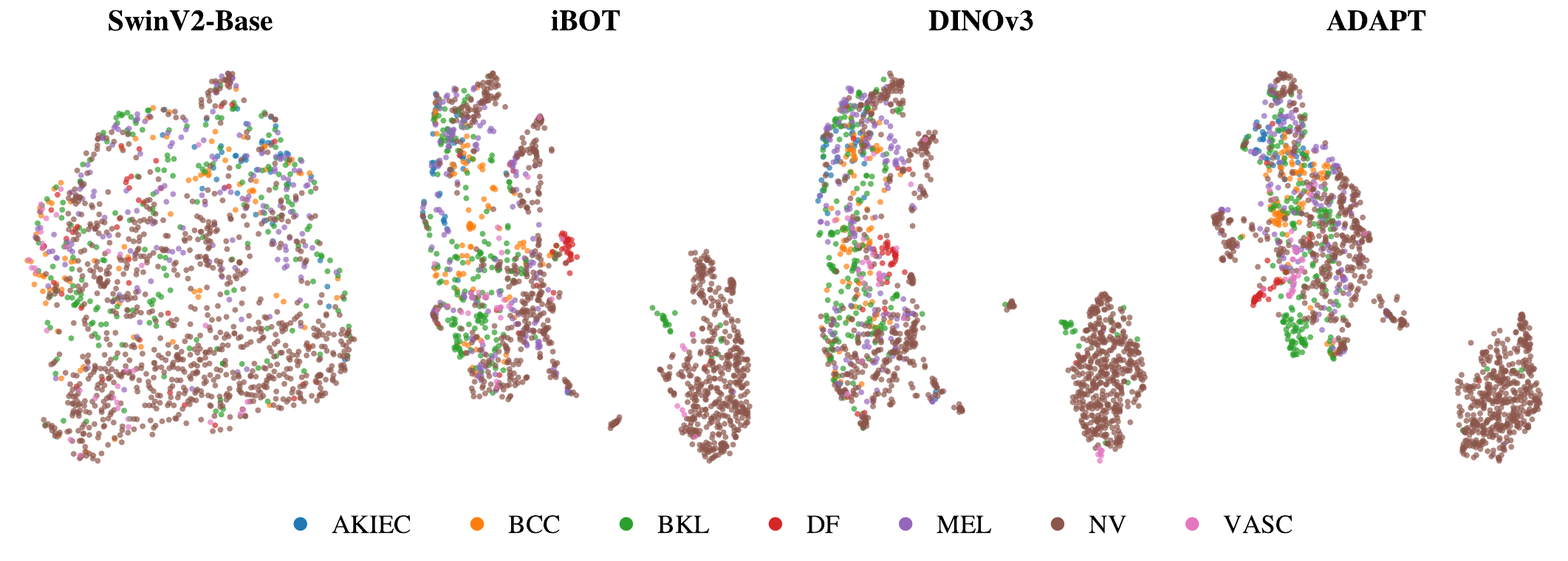}
\caption{UMAP of frozen encoder features on ISIC2018. AKIEC: actinic keratosis/Bowen's disease; BCC: basal cell carcinoma; BKL: benign keratosis; DF: dermatofibroma; MEL: melanoma; NV: melanocytic nevus; VASC: vascular lesion.}
\label{fig:umap}
\end{figure*}

With 10\% of the corpus, most of the gain over the ImageNet initialization (0.611) already comes from the global image-level objective (0.644); token-level targets add little (0.647). An unmasked local crop lowers the average back to the ImageNet level (0.612), and heavy masking is required: 25\% and 50\% masking reach only 0.623 and 0.620, whereas ADAPT's 75\% matches the global-only variants (0.648). Projection heads degrade linear probing (0.556 vs.\ 0.615), supporting the use of the teacher's backbone features as targets. Two global and eight local crops slow pretraining nearly 4$\times$ (6.1\,h vs.\ 1.6\,h) without improving accuracy (0.636).

As the global-only variants are within noise of ADAPT at this scale, we retrained them on the full corpus. Again, the global image-level objective accounts for most of the gain over the ImageNet initialization (0.585 $\rightarrow$ 0.641). ADAPT adds 0.005 (0.646) and is best under both protocols; the token-level loss does not help on its own (0.638) but does when combined with the masked local crop. This margin is within the seed standard deviation and costs 4.3\,h (about 40\%) more pretraining, yet ADAPT remains 3.0--4.4$\times$ faster than DINOv3 and iBOT (\cref{tab:efficiency}). When compute is limited, either global-only variant retains most of the classification gain at lower cost.

\subsection{Qualitative feature visualization}
To qualitatively assess the structure of the learned representation space, we visualize the frozen encoder features for the ISIC2018 dataset using UMAP~\cite{mcinnes2018umap} (\cref{fig:umap}). Features were extracted from the models trained with seed 42, and the UMAP projection was initialized with the same seed for reproducibility. 
The supervised ImageNet initialization (SwinV2-Base) shows strong class overlap within a single dense region. All self-supervised models produce more structured embeddings, but ADAPT achieves better class separation (especially grouping BCC \& BKL more compactly, and separating one NV subgroup more clearly). 
This is consistent with ADAPT’s linear probing results (\cref{sec:linear_probing}), confirming that ADAPT yields  discriminative representations.
\section{Discussion}
Our findings offer insights into visual representation learning under strong domain shift. General-purpose self-supervised frameworks such as DINOv3 provide strong baseline representations, but our results indicate that they leave room for domain-specific adaptation and incur substantial computational costs. ADAPT addresses both. With a streamlined pretraining design, it achieves the best average performance on downstream classification, outperforming its supervised ImageNet-initialized baseline as well as general-purpose and domain-specific self-supervised models such as DINOv3 and PanDerm, with the clearest gains on dermoscopic benchmarks. Its representations also transfer competitively to lesion segmentation. ADAPT outperforms the faster single-crop SimMIM on every benchmark while pretraining 3.0-4.4$\times$ faster than multi-crop baselines DINOv3 and iBOT.

\paragraph{Limitations and Future Work.} While our empirical evaluation focuses on dermatology due to its pronounced domain gap with natural images, the core principles of ADAPT are theoretically domain-agnostic. We hypothesize that this framework will generalize effectively to other medical imaging modalities, such as histopathology and radiology, as well as non-medical domains exhibiting severe visual divergence. Future work will also evaluate remaining local-crop parameters, such as optimal resolution and scale ranges, to further strengthen their contribution.
\section{Conclusion}
We introduced ADAPT, an efficient self-supervised domain adaptation framework explicitly designed to learn robust representations in the presence of severe visual domain shifts. By combining masked multi-crop learning with complementary token-level and image-level distillation objectives, ADAPT extracts both fine-grained local characteristics and global semantic features from unlabeled data.

Extensive evaluations on dermatology imaging, a challenging testbed for extreme domain adaptation, demonstrate ADAPT's superiority. Across multiple downstream benchmarks, ADAPT produced highly transferable representations, outperforming existing self-supervised baselines particularly in data-scarce dermoscopic regimes where expert annotations are limited. Furthermore, ADAPT achieves these state-of-the-art results with substantially lower computational overhead than standard multi-crop distillation frameworks.
{
    \small
    \bibliographystyle{ieeenat_fullname}
    \bibliography{main}
}

\end{document}